# PARAMETER-FREE MODEL OF RANK POLYSEMANTIC DISTRIBUTION
Victor Kromer

Russia, 630126, Novosibirsk, ul. Vilujskaja 28, NGPU

applied@nspu.nsu.ru

Novosibirsk State Pedagogical University

topic area – sociolinguistics

The purpose of this paper is the creation of a model of rank polysemantic distribution with a minimal number of fitting parameters. In an ideal case a parameter-free description of the dependence on the basis of one or several immediate features of the distribution is possible.

Probabilistic polysemantic distribution is peculiar to a word with a concrete frequency in the frequency list. Let's introduce dependence "usage - polysemanticism" proceeding from the following theoretical assumptions:
1. The lexicon of a concrete language (sublanguage) is reflected in the appropriate explanatory dictionary. A certain constitutive text corpus lays in the basis of the language.
2. Each use of a concrete word in the constitutive text corpus gives a new meaning.
3. A native language speaker does not distinguish separate meanings of the words and arranges them into groups of meanings (so-called "dictionary meanings") according to the psychophysical Weber-Fechner's law.

Let's make an assumption, that words in the constitutive text corpus are distributed in accordance with Zipf's law:

$$F = \frac{K}{i^\gamma}, \qquad (1)$$

where $F$ is the frequency of a word with rank $i$, $K$ is a constant of proportionality and $\gamma$ is Zipf's parameter for the distribution. According to Weber-Fechner's law, the spread of the process of arranging word-meanings onto groups of meanings, the number of dictionary meanings will be:

$$m_F = \psi(F+1) + C, \qquad (2)$$

where C=0.5772... is the Euler-Mascheroni constant.

Let's designate the number of words in the language considered as *L* and require equality of the rarest word in the constitutive text corpus, and accordingly the mathematical expectation of the number of its dictionary meanings, to unity. Let's introduce a normalization condition, that is we require equality of the sum of mathematical expectations of the number of word-meanings on all the ranks according to expression (2) and the total of meanings of all the words in the explanatory dictionary. We obtain a set of equations:

$$\begin{cases} \dfrac{K}{L^{\gamma}} = 1 \\ \sum_{i=1}^{L} \left[ \psi\left(\dfrac{K}{i^{\gamma}} + 1\right) + C \right] = M \end{cases} \quad (3)$$

Solving the set of equations (3), we find values of *K* and $\gamma$.

It is defined from experimental data that the probability of a word with a certain frequency to have *k* dictionary meanings is determined by the formula:

$$p_k = \frac{(m_F - 1)^{k-1}}{m_F^k}, \quad (4)$$

where $m_F$ is the mathematical expectation of the number of dictionary meanings for a word having frequency *F* in the constitutive text corpus. The number of words in the dictionary with *k* meanings will be:

$$N_k = \sum_{i=1}^{L} p_k = \sum_{i=1}^{L} \frac{(m_F - 1)^{k-1}}{m_F^k}, \quad (5)$$

where $m_F$ is determined according to formula (2), and *F* – according to formula (1) with parameters *K* and $\gamma$, defined from the set of equations (3) solution.

The model offered is tested on a series of the explanatory and author's dictionaries. The test was produced by comparison of the empirical and theoretical number of words having a certain degree of polysemy and evaluation of a goodness-of-fit test "Chi-square". For the evaluation of the criterion the words with a certain degree of polysemy were regarded as one class, and the small classes were integrated up to a size not less than 10 words. The least level of significance *P*, permitting us to reject the null hypothesis about equality of two compared distributions – the theoretical and empirical one was calculated. For the Pushkin's language dictionary *P*=0.21, and for the Dictionary of Russian in 4 volumes *P*=0.64. The null hypothesis about the correspondence of empirical and theoretical distributions can be accepted. For Ojegov's Dictionary of Russian *P*=0.03, and that formally allows us to reject the null hypothesis; however by surveying the empirical data the anomalous part of the polysemantic distribution for the words with polysemy degrees of 8 and 9 (inverse dependence) comes to light, and that can be probably explained by extralinguistic factors. When joining these two classes in one class *P*=0.51, which again allows us to accept the null hypothesis. For the Dictionary of Contemporary Russian in 17 volumes *P* is practically equal to zero, that is, the sharp distinction in compared distributions is observed. Accepting that the area of one-meaning words can fall out of the general tendency of the relations between volumes of word-groups of different polysemy degree, this means that it is possible as overfilling of peripheral area at the expense of engaging words concerning the lexical sphere outside the concrete dictionary type, as arbitrary exception of some dictionary words from this area, a conversion is possible of the parameter-free model into the one-parameter model with a fitting parameter *L\** – the number of words in the modified dictionary, at the expense of expansion or compression of the area of one-meaning words of the dictionary. Establishing the parameter *L\** as 157000 words, and by taking words with a number of meanings more than 14 out of consideration, *P* is equal to 0.03, and that allows us to accept the null hypothesis about equality of two distributions conditionally. There are about one hundred words with a degree of polysemy more than 14 in the dictionary under consideration, and they, as well as the words with one meaning, also fall out of the general tendency, described by the model offered.

So, the model offered allows us to describe polysemantic rank distributions of the lexicon of explanatory dictionaries on the basis of the observed distribution parameters – the total number of words and total number of meanings, and that means the model is a parameter-free one. In case of falling out of the area of one-meaning words from the general tendency the model requires introduction of one fitting parameter – the number of words in the modified dictionary.